%% file: boda-aa.tex
\newcommand{\arXiv}{} 

\ifdefined\anon
\newcommand{\bodaurl}{[withheld for blind review]} 
\else
\newcommand{\bodaurl}{https://github.com/moskewcz/boda} 
\fi

\ifdefined\nips
\documentclass{article}

\usepackage[final]{nips_2016}

\else
\documentclass[preprint,numbers]{sigplanconf}  
\newcommand{\sigplan}{} 
\fi

\def\vsp{\vspace{-0.05in}}
\usepackage[utf8]{inputenc} 
\usepackage[T1]{fontenc}    
\ifdefined\arXiv
\usepackage{hyperref}       
\else
\usepackage[draft]{hyperref}       
\fi

\usepackage{url}            
\usepackage{booktabs}       
\usepackage{amsfonts}       
\usepackage{nicefrac}       
\usepackage{microtype}      
\usepackage{color}
\usepackage{longtable}
\usepackage{appendix}

\usepackage[pdftex]{graphicx}

\usepackage{tabulary}

\usepackage{mathtools}
\usepackage{listings}
\DeclarePairedDelimiter\ceil{\lceil}{\rceil}

\newcommand{\dx}{{\times}}

\ifdefined\anon

\else

\fi

\title{A Metaprogramming and Autotuning Framework for Deploying Deep Learning Applications}

\ifdefined\anon
  \ifdefined\sigplan
    \authorinfo{Paper ID: \#158  , Number of pages: 12}{}{} 
  \else
    \author{ Authors withheld for blind review. } 
  \fi
\else
  \ifdefined\sigplan
  \authorinfo{Matthew W. Moskewicz \and Ali Jannesari \and Kurt Keutzer}
  {University of California, Berkeley}
  {\{moskewcz,jannesari,keutzer\}@eecs.berkeley.edu}
  \else
    \author{
      Matthew W. Moskewicz \qquad Ali Jannesari \qquad Kurt Keutzer\\
      University of California, Berkeley\\
      \texttt{\{moskewcz,jannesari,keutzer\}@eecs.berkeley.edu}\\
    }
  \fi
\fi

\begin{document}
\newcommand{\ndaii}[4]{\ensuremath{#1 \dx #2 = #3 \dx #4}}
\newcommand{\ndaiii}[6]{\ensuremath{#1 \dx #2 \dx #3 = #4 \dx #5 \dx #6}}
\newcommand{\ndaiiii}[8]{\ensuremath{#1 \dx #2 \dx #3 \dx #4 = #5 \dx #6 \dx #7 \dx #8}}


\maketitle

\ifdefined\arXiv
\thispagestyle{plain}
\pagestyle{plain}
\fi

\input{00_abstract}
\input{01_introduction}
\input{02_background}

\input{03_approach}
\input{06_evaluation}

\input{07_relatedworks}
\input{08_conclusion}


\ifdefined\anon
\else
\subsubsection*{Acknowledgments}
Research partially funded by DARPA Award Number HR0011-12-2-0016, the Berkeley Deep Drive (BDD) Industry Consortium, and ASPIRE industrial sponsors and affiliates Intel, Google, Hewlett-Packard, Huawei, LGE, Nvidia, Nokia, Oracle, and Samsung.
\fi

{
\small
\bibliographystyle{IEEEtran}
\bibliography{bibliography}
}

\newpage
\onecolumn
\input{09_appendix}

\end{document}

%% file: 00_abstract.tex
\begin{abstract}

  In recent years, deep neural networks (DNNs), have yielded strong results on a wide range of applications.
  Graphics Processing Units (GPUs) have been one key enabling factor leading to the current popularity of DNNs.
  However, despite increasing hardware flexibility and software programming toolchain maturity, high efficiency GPU programming remains difficult: it suffers from high complexity, low productivity, and low portability.
  GPU vendors such as NVIDIA have spent enormous effort to write special-purpose DNN libraries.
  However, on other hardware targets, especially mobile GPUs, such vendor libraries are not generally available.
  Thus, the development of portable, open, high-performance, energy-efficient GPU code for DNN operations would enable broader deployment of DNN-based algorithms.
  Toward this end, this work presents a framework to enable productive, high-efficiency GPU programming for DNN computations across hardware platforms and programming models. 
  In particular, the framework provides specific support for metaprogramming, autotuning, and DNN-tailored data types.
  Using our framework, we explore implementing DNN operations on three different hardware targets: NVIDIA, AMD, and Qualcomm GPUs.
  On NVIDIA GPUs, we show both portability between OpenCL and CUDA as well competitive performance compared to the vendor library.
  On Qualcomm GPUs, we show that our framework enables productive development of target-specific optimizations, and achieves reasonable absolute performance.
  Finally, On AMD GPUs, we show initial results that indicate our framework can yield reasonable performance on a new platform with minimal effort.

\end{abstract}

\keywords 
computer vision; code generation; neural networks; mobile computing; convolution

%% file: 01_introduction.tex
\section{Introduction}
\label{sec:intro}
\vsp
 
Modern Graphics Processing Units (GPUs) offer a tantalizing combination of general programmability, high peak operation throughput, and high energy efficiency.
However, despite increasing hardware flexibility and software programming toolchain maturity, high efficiency GPU programming remains difficult.
Only a small number of programmers can efficiently map new applications to GPUs, and even then the process often suffers from high complexity and low productivity.
Thus, in practice, the bulk of high-performance, high-efficiency GPU code resides inside highly tuned libraries for a few specific tasks.
Further, such libraries are generally tuned for only a small subset of GPUs -- typically only those from a single vendor.
Recently, interest in machine learning (ML) in general, and neural networks (NNs) in particular, has increased greatly.
Deep neural networks (DNNs) are emerging as the primary approach for challenging applications in computer vision, natural language processing (NLP), and human action recognition~\cite{ji20133d}.

As GPUs are well suited to (or perhaps more to the point, have enabled) current DNN approaches~\cite{alexnet},\cite{Schmidhuber2012}, much attention has been given to pushing the envelope of efficient GPU implementations for DNN computations.
Originally, GPU-based DNN computations leveraged existing dense linear algebra libraries (BLAS~\cite{cuBLAS},\cite{clblas}).
However, over several years, it became clear that more efficient special-purpose libraries were possible.
But, even given the high level of interest in the domain, and the significant speedups that were possible, the general availability of such libraries took years:
The landmark BLAS-based implementation from Krizhevsky~\cite{alexnet} was released in 2012-12.
However, the first official release of NVIDIA's cuDNN~\cite{cuDNN} library was not until 2014-09.
Over time, important improvements and new features have continued to appear, with cuDNN v5 released 2016-04.
These improvements and new features generally track new developments in the ML community, but with a significant real-time lag of many months.
Anecdotally, the reason for this long latency is simple: developing cuDNN requires a large amount of engineering effort by a team of specialized programmers, perhaps more than 15 staff-years to date.

In this work, we propose a framework for NN computations and aim to improve on the current state of high-efficiency special-purpose GPU programming in several ways.
First, we aim to shorten the time taken to prototype and tune high-efficiency GPU implementations of new algorithms and applications.
Second, we aim to improve the portability of such efforts across types of GPU hardware and programming models.
We achieve this with the careful application of both metaprogramming and autotuning in our proposed framework.
We demonstrate the approach via a case study of mapping DNN computations, particularly convolutions, to NVIDIA, Qualcomm, and AMD GPUs.
We show that we can target the same NVIDIA GPU hardware using either OpenCL or CUDA and achieve similar, high-efficiency results.
This portability is possible due to the metaprogramming support provided by the framework, which allows syntactic compatibility between the core languages of the programming models (i.e. OpenCL and CUDA).
Additionally, the framework abstracts away details related to compilation, allocation, scheduling, and execution that differ between OpenCL and CUDA.
Also, we show that our approach eases the cumulative burden of targeting NVIDIA, Qualcomm, and AMD GPUs.
Using the NVIDIA-tuned code as a starting point, we were able to achieve reasonable performance on a Qualcomm mobile GPU with only a few weeks of part-time effort by a programmer unfamiliar with the target.
Then, using all the code developed for the NVIDIA and Qualcomm platforms, we show the initial results of applying autotuning to target another new hardware platform, AMD GPUs.
With no manual effort, we achieve a modest level of performance on AMD hardware, and argue that the profiling and tuning capabilities of the framework provide a great starting point for platform-specific optimizations.

The rest of the paper is organized as follows. 
In Section~\ref{sec:background}, we review the key background concepts needed related to DNN computations, focusing particularly on convolution and ND-Arrays.
Then, in Section~\ref{sec:approach} we introduce our framework, the main contribution, with its different layers and structures.
We discuss DNN computations using Boda and describe the metaprogramming flow and autotuning methods.
In Section~\ref{sec:results}, we demonstrate our framework via case study and present the results.
Then, we discuss shortly the related works in Section~\ref{sec:relatedworks} considering the general topic of GPU programming and portability.
Finally, in Section~\ref{sec:conclusions} we conclude the paper.

%% file: 02_background.tex
\section{Background}
\label{sec:background}
\vsp

\begin{figure}[h]
	\centering
	\includegraphics[width=0.3\textwidth]{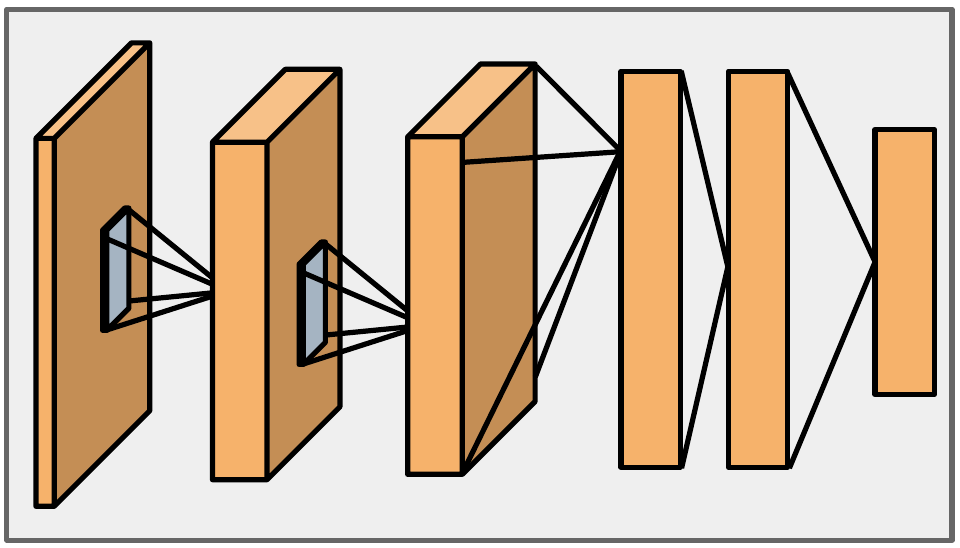}
	\caption{A neural network (NN) including convolution layers.}
	\label{fig:nn-layers}
\end{figure}

Deep neural networks (DNNs) have recently enhanced predictive power in many different machine learning applications.
Convolutional neural networks (CNNs) are DNNs which make heavy use of 2D \emph{convolutions} over multi-channel 2D images.
CNNs have been quite successful in many computer vision applications such as object detection ~\cite{girshick2015deformable} and video classification~\cite{karpathy2014large}.
The high-level idea behind CNNs is that they can learn a large set of image filters, organized into layers, where each layer feeds forward into the next.
Figure~\ref{fig:nn-layers} depicts a neural network that includes convolution layers.
Based on the intuition that useful features for vision should be spatially invariant, the learned filters are applied uniformly at many points to their input images, using a sliding window approach.
Each filter in each layer thus defines a feature which is itself a single-channel image.
The outputs of all the filters from each layer are then stacked to form the multi-channel image output for that layer.
As layers proceed from input to output, the features are increasingly-high-level abstractions of the input image content.
For example, the initial layers of a CNN might learn features indicating corners, edges, and textures.
Intermediate layers might combine these to form higher-level features such as ``fabric-ness'', ``cat-eyes'', and so forth.
Lastly, the final layer produces task-specified features such as ``cat-in-image'', ``person-in-image'', and so forth.

In addition to \emph{convolutions}, CNNs commonly contain other operations such as \emph{pooling} and \emph{nonlinear activation functions}.
However, for CNNs, convolution operations typically dominate the computation.
Typically, convolutions require many operations (100s to 1000s or more) to produce a single output pixel, as each output pixel depends on all input pixels across all input channels within a convolution kernel-sized window of the input.
In contrast, activation functions typically are applied element-wise, and require only one or a few operations per output pixel.
Similarly, the most common type of pooling, spatial max-pooling, typically has a small window (often $3 \dx 3$) and operates per-channel, thus only requiring few ($\sim 9$ for the $3 \dx 3$ case) operations per output pixel.
Note that, while activation functions and pooling may require little computation, they still may incur significant memory access overhead and/or require some care to implement efficiently.

\begin{figure}[t]
	\centering
	\includegraphics[width=0.24\textwidth]{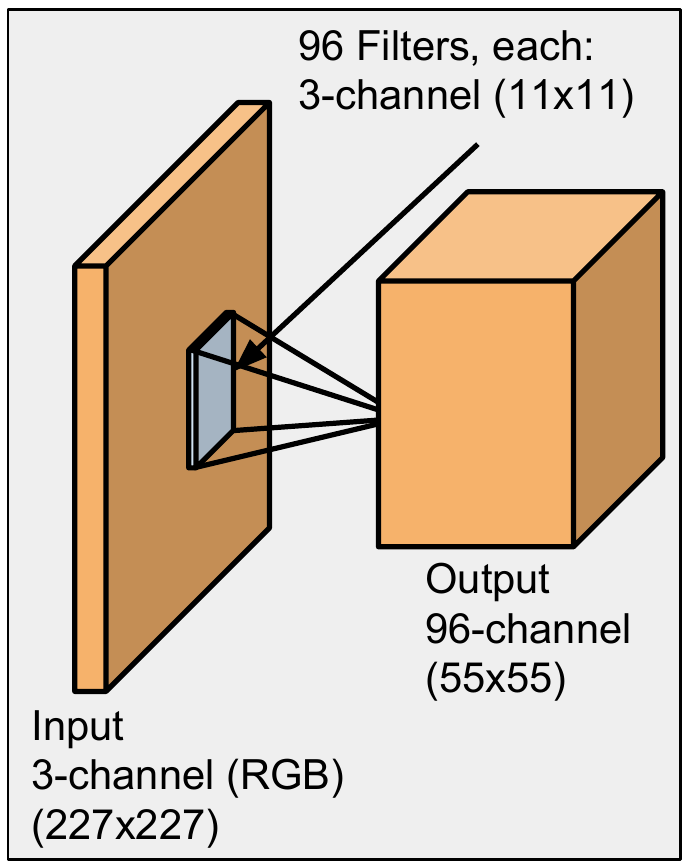}
	\caption{A single convolution -- dot-product: a filter as sliding window function applied to a matrix.}
	\label{fig:dot-prod}
\end{figure}

\emph{ND-Arrays}, or collections of numbers with N indexes (sometimes also called N-D Matrices or tensors), are the main data type used for CNN computations.
In particular, the input image, the filtered images produced by each layer (and fed as input to the next layer), and the filters themselves are all ND-Arrays.
That is, each layer of convolutions in a CNN can be defined as the function \emph{output = conv (input, filters)}, where \emph{output}, \emph{input} and \emph{filters} are all ND-Arrays.
Figure~\ref{fig:dot-prod} shows an example of a single convolutional layer with 96 filters applied to an input of a single multi-channel (RGB) image with size $227 \dx 227 \dx 3-channel$.
Each filter has size $11 \dx 11 \dx 3-channel$, and is slid over the input with a spatial stride of 4.
Thus, \emph{output} has size $55 \dx 55 \dx 96-channel$.
Each output value of the convolution is the result of a dot-product between a filter's weights and an $11 \dx 11$ region of the input image.
In the remainder of this work, we will predominantly focus on the efficient implementation of such convolution operations.

%% file: 03_approach.tex
\section{Approach}
\label{sec:approach}
\vsp

\begin{figure}[h]
	\centering
	\includegraphics[width=0.49\textwidth]{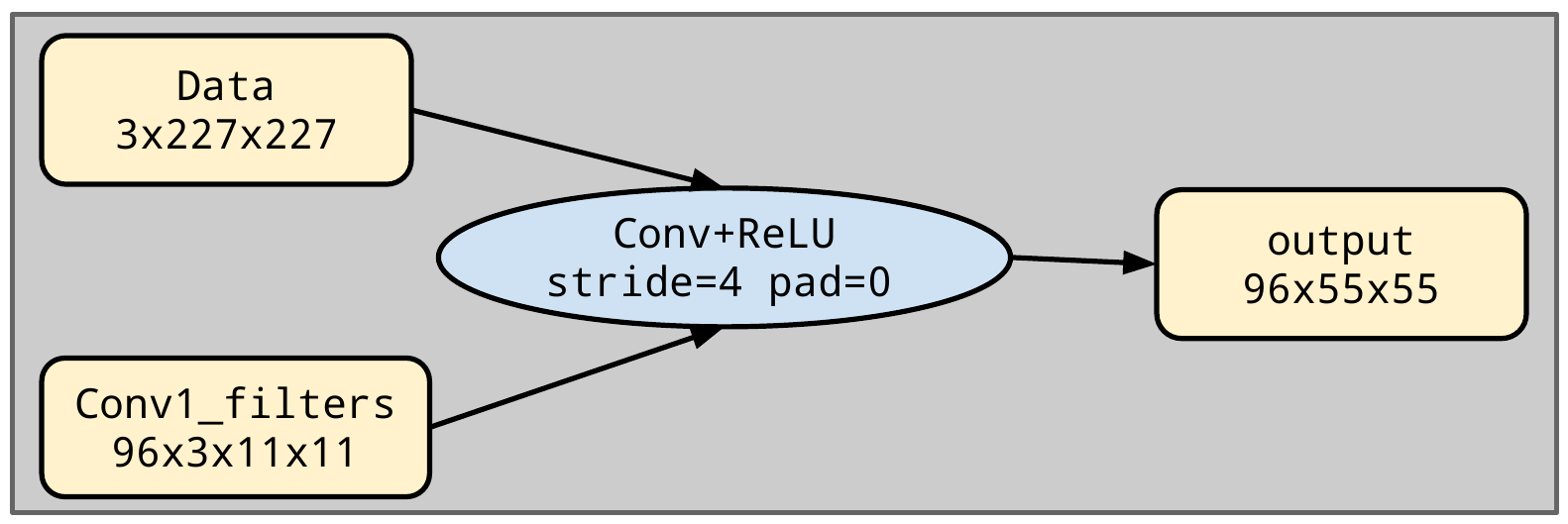}
	\caption{The compute graph of the convolution function in Figure~\ref{fig:dot-prod}.}
	\label{fig:compute-graph-example}
\end{figure}

We are motivated by the desire to efficiently deploy computationally intensive NN-based applications across various hardware platforms.
Thus, our main task is simply expressed: given a NN and its inputs, efficiently compute its outputs.
At a high level, a NN is simply a (stateless) function from its inputs to its outputs.
We can define a NN as a directed acyclic graph of primitive operations and ND-Arrays.
Figure~\ref{fig:compute-graph-example} demonstrates the compute graph of the single convolution function shown in Figure~\ref{fig:dot-prod}.
We refer to this type of graph as a \emph{compute graph}.


So, we can restate our task more generally as:
Given a NN's compute graph and values for its input (source) ND-Arrays, compute all of its output (sink) ND-Arrays.
But, our task is simplified by a few key properties of NN compute graphs:
\begin{itemize}
\item The set of operations we must support is relatively small ($\sim$10 unique operations), with only one (Convolution) being computationally critical.
\item The size of the graphs are generally small ($\sim$10-100 nodes), and generally each operation node represents a significant amount of work. Thus, node evaluation order (i.e. scheduling) is typically not critical.
\item There are only a few important graph-level (cross-operation) optimizations, and these are both local and simple (typically pair-wise fusions).
\end{itemize}

In practice, NNs are typically either:
\begin{itemize}
\item Explicitly specified using some concrete graph syntax (e.g. a list of nodes and edges such as the Caffe~\cite{jia2014caffe} .prototxt format)
\item The output of a graph-generator program (itself typically written in a high-level language such as Python or Lua~\cite{Lua}).
\end{itemize}

We view both of these methods as Domain Specific Languages (DSLs) for NN specification.
We term the conversion from DSL to compute graph as the \emph{NN front-end}.

\subsection{Front-end Layer}
In this work, we are NN front-end neutral; as long as a suitable compute graph can be produced, any NN front-end can be used with our approach.
Currently, our framework provides only a reader for the Caffe .prototxt format, but this is sufficient for all our experiments.
However, adding support for additional NN front-ends is a straightforward task (using the existing reader as a model).




% ...

Once we have a compute graph, we turn our attention to the operations that are used inside it.
Typically, the individual operations inside the compute graph are specified by name (e.g. ``Convolution'' or ``Pooling'').
Additionally, each operation may have various configuration parameters (e.g. ``stride=2'') which alter its semantics.
It is very typical that a given NN will use mostly common operations plus one or a few uncommon or unique operations.
For example, the ``Deconvnets'' work~\cite{zeiler2014visualizing} introduced the Deconvolution operation.
Similarly, work on object detection~\cite{yu2015multi} used dilated convolution, which generalized the existing convolution operation by adding a dilation parameter.
In both works, all other NN operations used were common ones.
Thus, regardless of what NN front-end(s) are supported, it is the support (or lack thereof) for individual operations and parameter settings that limits what applications can be supported at the compute graph layer.
As mentioned above, in this work we mainly focus on support for the \emph{Convolution} operations which dominate the computation, and then only on commonly used parameter settings.
However, we additionally provide enough support for ancillary functions to handle a sampling of common full networks.
Still, we recognize that the ability to add support for new operations is also an important issue.
Based on our overall experience adding support for the various needed ancillary operations, we claim that adding additional operations to our framework is generally straightforward.
Often, such new operations are neither particularly complex nor do they require significant computation.
So, the challenge with such operations becomes that of either:
\begin{itemize}
\item Producing reasonable quality (but not necessarily highly-tuned) framework-native implementations, or
\item Providing suitable wrappers and/or glue to integrate existing implementations.
\end{itemize}

\subsection{Boda Overview -- Branching Stovepipe}
\begin{figure}[h]
	\centering
	\includegraphics[width=0.49\textwidth]{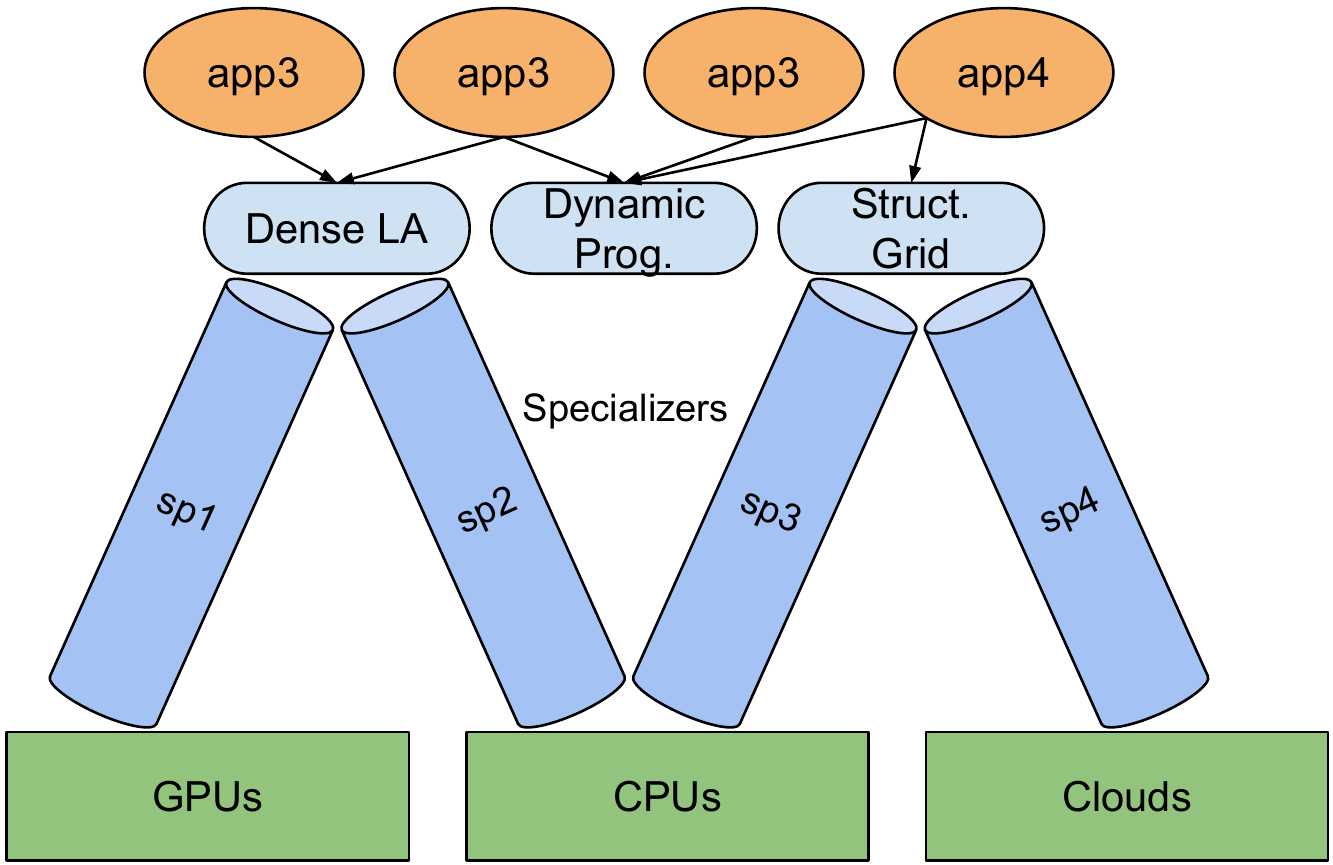}
	\caption{Specializers as stovepipes (one stovepipe per target).}
	\label{fig:specializers-stovepipes}
\end{figure}

Our framework is inspired  by the SEJITS~\cite{catanzaro2009sejits} methodology.
However, since our chosen top-level domain-specific language (DSL) is a simple compute graph (or a program that generates a compute graph) of known operations, we avoid much of the front-end complexity typically present in SEJITS flows~\cite{truong2016latte}.
The intent of our framework is to provide a direct, complete path from NN DSLs to execution.
In particular, we trade-off front-end generality for having only a small, known set of computationally-critical operations we must handle.
We began our exploration of this idea, focused on a single target and programming model, in our prior work~\cite{moskewicz2016boda}.
Here, we extend that work to fully explore multiple hardware targets and programming models.
We embrace the idea of the ``stovepipe'': providing a special-purpose flow that maps a single DSL to a hardware target as shown in Figure \ref{fig:specializers-stovepipes}.

\begin{figure}[h]
	\centering
	\includegraphics[width=0.49\textwidth]{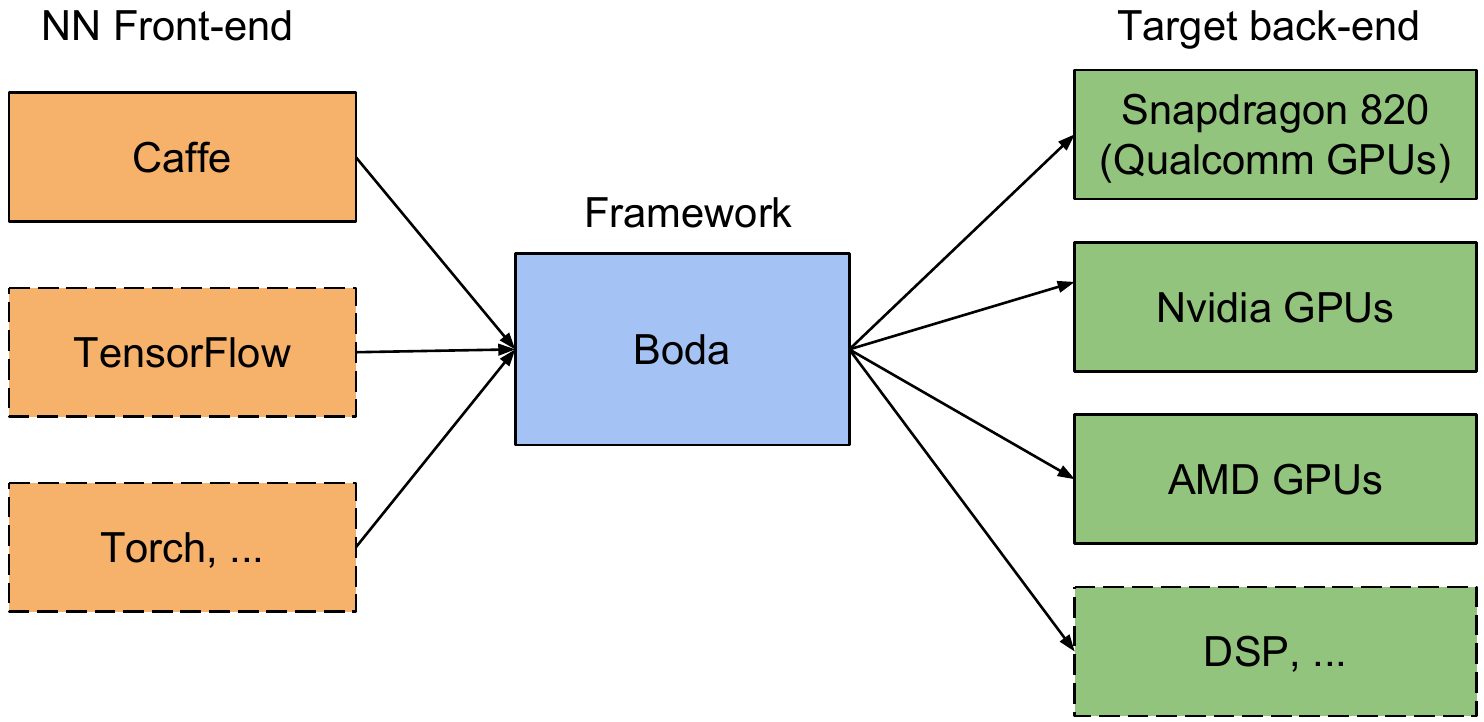}
	\caption{Boda framework as branching stovepipes (specializers) providing maximum efficiency and portability.}
	\label{fig:boda-stovepipes-overview}
\end{figure}

Such an approach is a middle ground between the traditional library and compiler approaches to the mapping problem:
\begin{itemize}
\item Compared to a traditional library, our approach is more complex but much more flexible.
\item Compared to general-purpose compilation, our approach is more limited but much simpler.
\end{itemize}
In particular, compared to a full compiler, we avoid \emph{all} mandatory, pre-existing, already-specified intermediary layers present in typical compilation flows.
But, being somewhat compiler-like compared to the traditional library approach, our approach is well suited to exploit metaprogramming techniques.
Overall, using this approach, we can achieve high efficiency for a limited set of operations with manageable complexity.
One key to reduced complexity is that, at runtime, we need only handle the specific operations present in the input.
Thus, we need not attempt to create or compile code for all possible cases as a library would.
Further, unlike a library, we are free to use \emph{all} input-specific information to aid in optimization.
In particular, for each operation, we need only handle the specific input sizes used.
As the number of possible input sizes is very large, such specialization is cumbersome and/or limited in the traditional library approach.
Traditionally, the stovepipe approach targets only a single hardware platform for execution.
With its focus on a single, specific mapping task, the stovepipe approach offers a good combination of freedom, flexibility, and simplicity.
However, as we desire \emph{a limited degree} of portability to multiple similar hardware targets, we compromise somewhat and extend the approach.
We term our extended approach \emph{branching stovepipes}: rather than one stovepipe per target, we can share much effort across related targets, at the expense of some added complexity.
Figure \ref{fig:boda-stovepipes-overview} depicts the overall idea.
Thus, a careful balance between portability, complexity, and efficiency is a key design goal of our framework.

\subsection{Framework Structure}
\begin{figure}[h]
	\centering
	\includegraphics[width=0.49\textwidth]{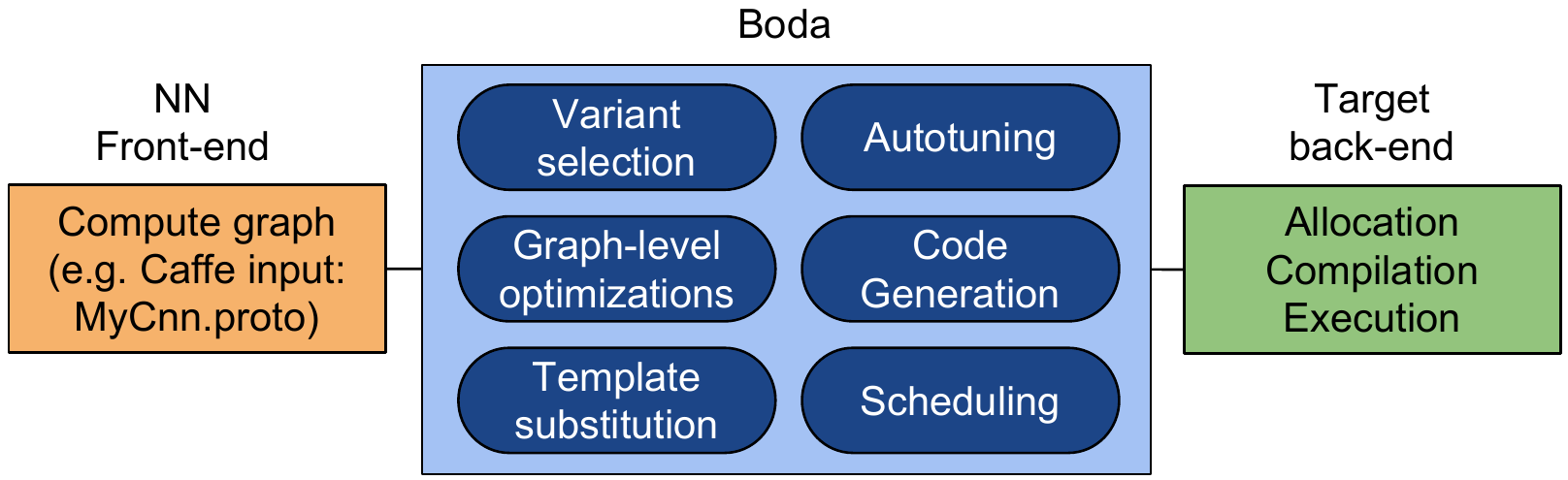}
	\caption{The layering and structure of Boda framework.}
	\label{fig:boda-structure}
\end{figure}

The exact layering and structure of our approach is shown in Figure~\ref{fig:boda-structure}.
The structure is extensible and flexible to evolve to meet changing needs.
Bearing this in mind, we now give an outline of the current layers/stages in our framework for mapping NN computations to GPU hardware.


We have already discussed first layer of our framework, the NN front-end layer, which converts NN DSLs to compute graphs.
Now, we will discuss the \emph{target back-end} layer where we hand off responsibility to pre-existing target-specific programming/development platforms.
Generally, as with the front-end, our framework is back-end neutral.
We require only that the target platform provide mechanisms for:
\begin{itemize}
\item Memory allocation,
\item Compilation and execution of code, and
\item Per-function-invocation execution timing.
\end{itemize}
We make no attempt to achieve portability beyond our target set of back-ends.
Thus, we need not support arbitrary languages or programming models throughout our framework, but only what is necessary for the back-ends we wish to target.
Conveniently, all modern GPUs support similar programming models and input languages.
NVidia hardware supports both CUDA~\cite{CUDA} and OpenCL~\cite{stone2010opencl}.
Other hardware vendors, such as AMD and Qualcomm, support only OpenCL.
Both OpenCL and CUDA offer comparable interfaces for memory allocation, compilation, and execution of code.
Further, the core language for describing computation supported by both OpenCL and CUDA has C-like syntax and semantics.

\paragraph{Programming model portability with CUCL.}
While there are syntactic differences and various language (or hardware) specific features, the core semantics and programming models of OpenCL and CUDA are fairly compatible.
Thus, across the two platforms, we can share varying amounts of our framework's metaprogramming support, per-operation metacode, and per operation code templates.
We term \emph{CUCL} as the intersection language formed from the cross-compatible subset of CUDA and OpenCL.
We use a small amount of text substitution to bridge syntactic differences (e.g. CUDA's \texttt{threadIdx} versus OpenCL's \texttt{get\_local\_id}).
When necessary or desired, though, our framework fully allows the use of back-end specific languages or features.
As one would expect, use of such back-end specific features limits portability.
Yet, a far more limiting and important issue is that of \emph{performance portability}.
While it is convenient to share a common syntax and semantics for computation (i.e. C) across targets, this ensures only functional equivalence.
In particular, this functional equivalence is very helpful for development, testing, and debugging.
However, it does not address our goal of achieving high efficiency across all back-ends.
We make no attempt to create or rely upon a general-purpose compiler to achieve performance portability.
Instead, we aim to minimize the effort needed in order to specialize our few operations of interest across our limited set of target back-ends.

\paragraph{Performance portability with metaprogramming.}
Together, the NN front-end layer and per-hardware-target back-end layers define the interface and scope of our framework.
The remaining layers form the \emph{branching stovepipe} which connects the front-end to the back-ends.
That is, these layers form a self-contained metaprogramming-based flow designed to execute NN compute graphs.
As discussed earlier, the high peak compute rates of GPUs make them attractive targets for NN convolutions.
Further, it is clear that reasonably high efficiency has been achieved in many instances, albeit with significant programmer effort.
In particular, evidence suggests that such efforts generally involve both low level programming and a significant degree of metaprogramming ~\cite{maxDNN} ~\cite{cuDNN}.
Similarly, even if using existing BLAS libraries as building blocks, evidence suggests that usage of metaprogramming in libraries such as cuBLAS~\cite{cuBLAS}, MAGMA~\cite{magma-autotuning}, and clBLAS~\cite{clblas} is also commonplace.
Thus, the novelty of our approach is not merely the usage of metaprogramming, but in the specific design choices made to balance efficiency, portability, complexity, and usability.
We base our design decisions on the specific goals, constraints, and challenges of the mapping to be performed.

\paragraph{ND-Arrays.} Our first guiding observation is that ND-Arrays are central to NN operations, as the majority of operation inputs and outputs are ND-Arrays.
Hence, ND-Array specific support for metaprogramming forms a cornerstone of our approach.
ND-Arrays are quite simple data structures.
Typically, they follow the format of standard C-style multidimensional arrays:
A single contiguous block of memory filled with a single flat array of elements.
Importantly, in our application, the sizes of all such arrays are known and fixed at the compute graph level.
Thus, we may \emph{statically specialize all operations} based on the sizes of their input and output ND-Arrays.
All indexing and bounding operations on such ND-Arrays may be reduced to chains of constant multiplies, divides, and modulo operations.
The resultant expressions are then amenable to both metaprogramming-level and low-level optimizations.
Further, in user-written templates, we require that all dimensions of each ND-Array must be named.
This use of mnemonic, semantically-significant names for array dimensions helps clarify code using ND-Arrays.
By analogy, imagine code that used C structures where each field was simply referred to by index rather than name.
Not only do named ND-Array dimensions improve readability, but they are used to implement a form of type checking for all ND-Array arguments.
All ND-Array arguments passed to a function must have the same number of dimensions \emph{with the same names} as given in their argument declarations.
For example, a function expecting a 4D-Array with dimension names \emph{in\_chan:out\_chan:y:x} (i.e. a set of filters) could not be passed a 4D-array with dimension names \emph{img:chan:y:x} (i.e. a batch of images).

\subsection{Overall Metaprogramming Flow}
\begin{figure}[h]
	\centering
	\includegraphics[width=0.49\textwidth]{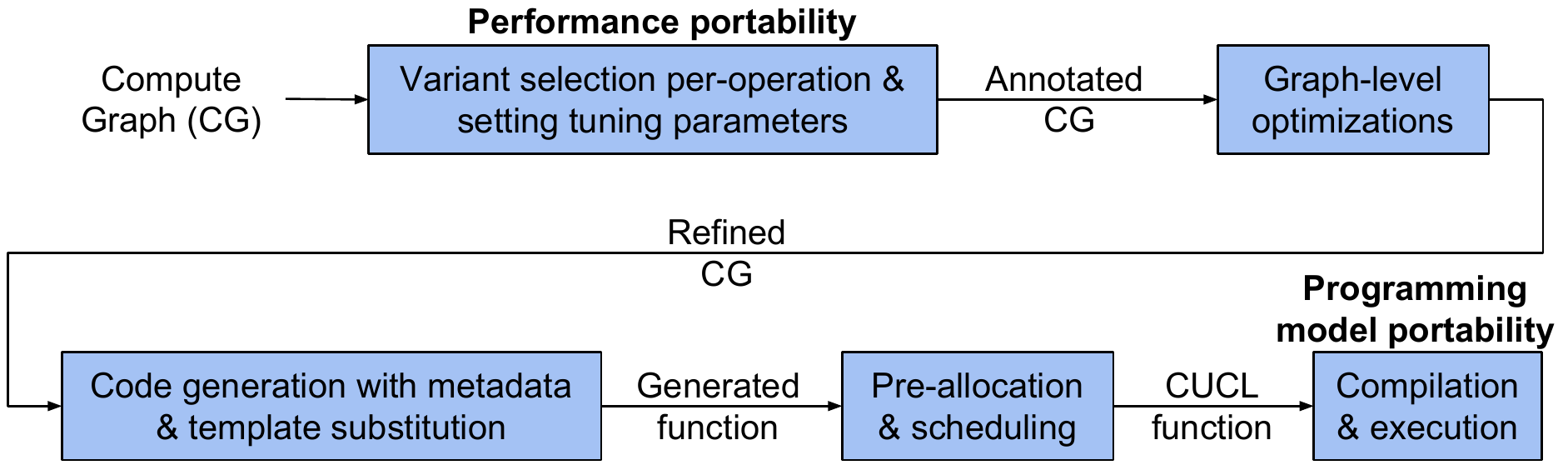}
	\caption{Boda flow: from compute graph to code.}
	\label{fig:boda-flow}
\end{figure}

Now, we discuss our overall meta-programming flow, which includes the framework layers shown in Figure \ref{fig:boda-flow}.
As discussed earlier, metaprogramming is, by necessity, commonly used to create high efficiency GPU implementations of NN operations.
Thus, the challenge of our framework is how best to support and guide the usage of metaprogramming.
We start with allowing the user to write only mildly restricted native GPU code in our CUDA/OpenCL subset language, CUCL.
Compared to directly using CUDA or OpenCL, CUCL:
\begin{itemize}
\item provides language-neutral idioms to replace those from CUDA and OpenCL, and
\item requires all ND-Array function arguments to be decorated with their dimension names, and
\item requires access to ND-Array metadata (sizes, strides) to use a special template syntax: $\%(myarray\_mydim\_size)$.
\end{itemize}
Many simpler operations can be directly written as a single \emph{CUCL function template}.
To produce OpenCL or CUDA functions from a CUCL function template, the framework: (1) replaces CUCL idioms with OpenCL or CUDA ones, and (2) replaces references to ND-Array sizes and strides with either (at the user's explicit choice) (2a) constants for the specific input ND-Array sizes, or (2b) references to dynamically-passed ND-Array metadata.
Typically, we care most about the case where the sizes are replaced with constants, as this gives the most possibility for optimizations and therefor efficiency.
However, this does require instantiation of the given CUCL template for \emph{every} unique set of called argument sizes.
Sometimes, for a given operation, this is unnecessary for performance, and perhaps even creates prohibitive overhead.
Thus, our framework also allows for the ``normal'' approach of simply passing the sizes and strides of ND-Arrays dynamically as function arguments.




\paragraph{Metaprogramming for NN Convolutions.}
NN convolution can be viewed as generalized matrix-matrix multiplication.
In fact, in the early era of NN computation, NN convolution was often implemented using BLAS (Basic Linear Algebra Subroutines) library SGEMM (Single-precision General Matrix-Matrix multiply) calls for the bulk of the computation.
But, as discussed in Section~\ref{sec:relatedworks}, the use of special-purpose code for NN convolutions is currently the dominant approach.
However, writing efficient NN Convolution code is difficult, as it requires:
\begin{itemize}
\item many large blocks consisting of many moves or multiplies, and
\item supporting many regimes of input sizes and convolution parameters, and
\item having a high degree of control over data movement and execution.
\end{itemize}
All of these issues share a common solution: metaprogramming.
With metaprogramming, one can easily write loops at the metacode level to generate long sequences of moves or multiplies.
Further, one can write multiple code generators to handle different input regimes, and select between them with metacode level case-splits.
Finally, one can generate memory and register indexing code and patterns without repetitive, error-prone manual effort.
Prior efforts have indeed uniformly used metaprogramming to varying degrees to address these issues; see Section~\ref{sec:relatedworks} for more discussion and a detailed comparison with this work.
At a high level, we choose to take a very general and flexible approach to metaprogramming.
Rather that use some language-level metaprogramming facility, we choose to directly write code generators in our framework's host language of C++.
We use our framework's native support for ND-Arrays at the metacode layer to (when desired) allow code generation to exploit fixed, exact sizes for all inputs and outputs.
For example, when cooperatively loading data across threads on a GPU, one must typically employ a loop with a conditional load.
If there are $N$ threads loading $W$ words, the loop must iterate $\ceil{W/N}$ times.
For each iteration, the load must be guarded on the condition that $i*N+thread\_id < W$.
In CUCL, OpenCL, or CUDA, here is a simplified version of how such a loop might appear:
\lstset{basicstyle=\ttfamily\scriptsize}
\begin{lstlisting}[language=C]
  for(int i = 0; i < ((W-1)/N)+1; ++i) {
    int const ix = i*N + thread_id;
    if(ix< W){filts_buf[ix] = filts[ix];}
  }
\end{lstlisting}
However, if $N$ and $W$ are fixed, we know we need exactly $\ceil{W/N}$ individual loads.
Further, only the last load need be conditional, and then only if $(W\bmod N)$ is non-zero.
In some cases, just making W and N constant may allow the platform-specific compiler to unroll the loop and eliminate unneeded conditionals without additional effort.
We show our framework's support for this simple metaprogramming approach here, where we have replaced the W and N variables with template variables that will be expanded to integer string constants:
\begin{lstlisting}[language=C]
  #pragma unroll
  for(int i = 0; i < ((%(W)-1)/%(N))+1; ++i) {
    int const ix = i*N + thread_id;
    if(ix<%(W)){filts_buf[ix] = filts[ix];}
  }
\end{lstlisting}
Although simpler metaprogramming approaches (such as C++ templates) might be sufficient to handle this case, we have observed that the platform-specific compiler often does not successfully unroll the loop and remove unneeded conditionals.
In such cases, our framework allows us to smoothly and easily shift more complexity to the metacode level and directly emit the sequence of desired loads.
To do this, we move the loop to the metacode level, and replace it entirely with a template variable in the CUCL code:
\begin{lstlisting}[language=C]
  %(filts_buf_loads);
\end{lstlisting} 
Then, at the metacode level, we write code to generate the needed sequence of loads, which is similar in structure to the original loop:
\begin{lstlisting}[language=C]
  string ix_str, load_str;
  for(int i = 0; i < ((W-1)/N)+1; ++i) {
    int const max_ix = i*N + (N-1);
    ix_str = str(i*N)+"+thread_id";
    load_str  = "filts_buf["+ix_str+"]";
    load_str += "= filts["+ix_str+"];";
    if(max_ix >= W){ // need bound check
      load_str = "if("+ix_str+"<"+str(W)
               + "){"+load_str+"}";
    }
    emit( "filts_buf_loads", load_str );
  }
\end{lstlisting}
When this metacode is run for a specific case, (here we show N=96,W=256) and the CUCL template is instantiated, the result is exactly the desired sequence of loads:
\begin{lstlisting}[language=C]
  filts_buf[0+thread_id] = filts[thread_id];
  filts_buf[96+thread_id] = filts[96+thread_id];
  if(192+thread_id<256){
    filts_buf[192+thread_id] = filts[192+thread_id];
  }
\end{lstlisting} 

In an observed, important case (with N=64,W=256), this metacode approach emitted code that resulted in 4 binary-level load instructions.
In contrast, all the loop-based versions emitted dozens of instructions including several conditionals, and more importantly, were much slower to execute in practice.
Similar issues arise in the generation of other memory accesses and also the generation of the fused multiply-adds that are the core computations needed for convolution.
Although it is not necessarily easy to find or determine what sequences of C-level code will execute well on a given platform, our framework aims to make the process easier.
In general, making as much as possible in the code constant, while reducing the usage of loops and conditionals, increases the chance that the platform-specific compiler can generate efficient binary code.
Further, the ability to easily profile different compute and memory access patterns by making changes at the metacode level, without needing to manually rewrite large sections of target-specific code, provides a large productivity boost.

\subsection{Variant Selection and Autotuning}
As mentioned, NN Convolutions have a wide range of possible input sizes and parameters.
It is generally difficult, even with metaprogramming, to write code that runs well over more than a limited range of input sizes.
Furthermore, many algorithms require target-dependant approaches that maybe be difficult or unfavorable to unify into a single function.
Thus, depending on which cases are important, we expect the user will create multiple \emph{variants} of each operation.
Further, each \emph{variant} may have various \emph{tuning} parameters that affect code generation.
Such tuning parameters might control thread blocking, memory access patterns, or load/store/compute vector widths.
Consider a typical set of tuning parameters and their values: \emph{MNt=4:4,MNb=16:16,Kb=4,vw=4}.
These parameters specify $4 \dx 4$ register blocking, $16 \dx 16$ thread blocking, an inner-loop-unroll-factor of 4, and a vector/SIMD width of 4.
Given an input size and target platform, it may be tractable to manually or heuristically choose a variant and its tuning parameters -- particularly when variants are written with specific targets and input size ranges in mind.
However, when considering many operations across many input sizes across many target platforms, this task becomes at best onerous and at worst intractable.
Thus, an important complimentary technique is \emph{autotuning}, where such parameters can be selected automatically by the framework.
By performing a brute-force, guided, or sampled exploration of the space of variants and tuning parameters, we can both: (1) find the best parameters for a given operation, as well as (2) learn much about a new target platform.

\begin{figure}[h]
	\centering
	\includegraphics[width=0.49\textwidth]{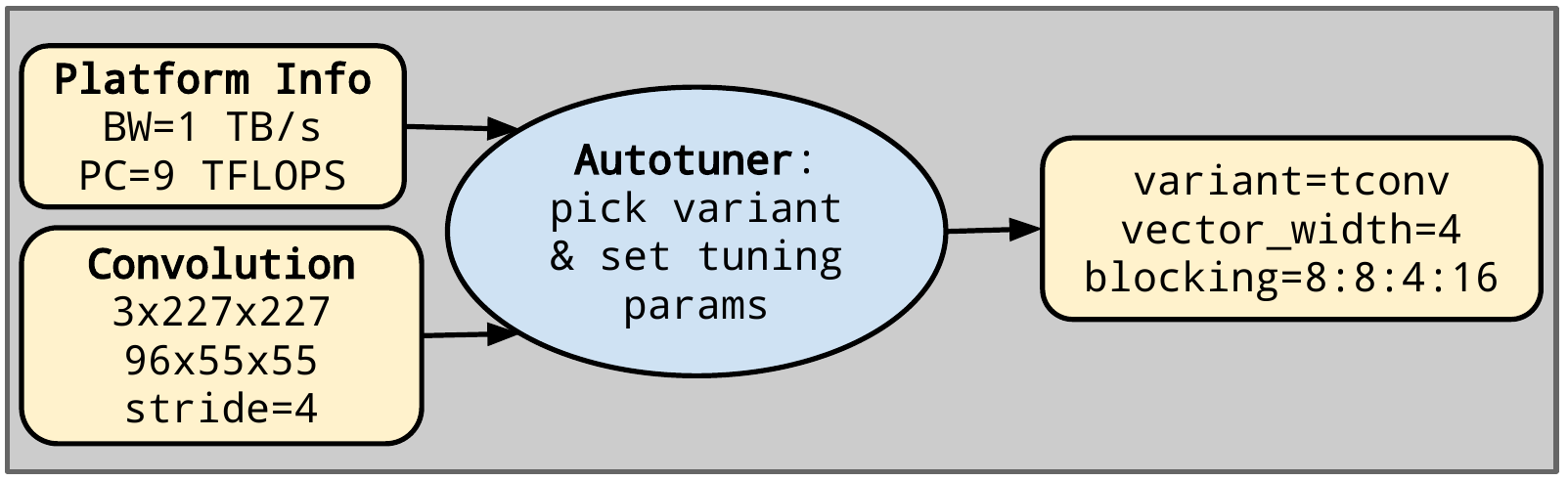}
	\caption{Boda autotuning.}
	\label{fig:autotuning}
\end{figure}

Figure~\ref{fig:autotuning} demonstrates the key features of autotuning: automatic per-platform variant selection and automated sweeps over tuning parameters.
Currently, we apply a simple brute-force search over a fixed set of configurations. 


\subsection{Graph-level Optimizations}
Next, we discuss graph-level optimizations: a critical but relatively simple part of our flow.
In particular, there are two important graph-level optimizations for NN compute graphs:
\begin{itemize}
\item Fusing of adjacent convolution and activation operations, and
\item Inserting any needed data-format-conversion operations.
\end{itemize}
The fusion optimization is straight-forward.
Convolution operations are commonly followed by the application of some element-wise activation functions.
In some cases, the overhead to read and re-write the output ND-Array to apply the activation function is significant.
In these cases, one may inline the code for the activation function in the convolution operation to avoid a read-modify-write of the output.
While this may significantly increase the code size of the output-writing part of the convolution operation, it is generally still favorable to do this.
So, our framework simply always performs this fusion when possible, using string substitution to insert an application of the activation function for all output-value writes.
The insertion of data-format-conversion operations is necessary due to the fact that some variants may use different layouts or padding of their input or output ND-Arrays.
That is, since we are able to freely choose the format of most internal ND-Arrays, we can exploit this to achieve higher efficiency within each variant.
However, after variant selection is complete, we must ensure that all ND-Arrays are transformed into the proper formats prior to each operation, and thus the potential need to insert data-format-conversion operations.



\subsection{Code Generation, Scheduling, and Execution}
As previously described, the final CUCL templates needed to cover all operation nodes in the compute graph are instantiated into OpenCL or CUDA and then compiled.
Then, once we have generated and compiled callable functions for each operation, we execute the compute graph.
For this, we must first consider operation scheduling and ND-Array allocation.
For our current target application, scheduling is not difficult.
The bulk of execution time is spent on operations that can each individually saturate the target hardware's compute capacity by themselves.
So, we need not attempt to run multiple operations in parallel; any topological sort of the compute graph yields a reasonable execution order.
Further, for our current use cases, we are generally not limited by GPU memory.
Hence, we can employ a naive allocation strategy and simply pre-allocate all ND-Arrays in the compute graph.
However, with some additional work, our framework should be easily capable of supporting more complex scheduling and allocation policies if needed or desired.
After allocation and scheduling, we issue the resultant sequence of function calls to the target back-end, which in turn performs all the desired computations.
The output ND-Arrays are then resident in GPU memory, ready to be read back to the CPU or processed further as desired.


%% file: 06_evaluation.tex
\section{Results}
\label{sec:results}
\vsp
We now report per-convolution-operation runtime results across hardware targets and programming models, organized to illustrate the key contributions of our work.
\paragraph{Experimental Setup.}
The benchmark set of convolution operations was chosen by extracting the unique convolutions from three common DNNs: ``AlexNet''~\cite{alexnet}, ``Network-in-Network''~\cite{Lin2013NiN}, and the first version of Google's ``Inception'' network~\cite{googlenet}.
For reporting results here, we choose a relatively small, but representative, set of operations\footnote{The full set of operations with all parameters and results is available online at \bodaurl\  -- along with our framework and all needed tools to replicate these results and figures.}.
In particular, we include only operations with:
\begin{itemize}
\item a batch size of 5, which models a streaming deployment scenario with some latency tolerance, and
\item have more than $1e8$ FLOPS (as we focused our optimization effort more toward these sizes).
\end{itemize}
We organize the operations by sorting them by FLOP count, which is a reasonable proxy for the difficulty of a given operation.
However, depending on the exact convolution parameters, two operations with similar FLOP counts may substantially differ in both:
\begin{itemize}
\item their theoretical maximum efficiency for a given hardware platform (based on Roofline analysis), as well as
\item the empirical performance of any given convolution algorithm.
\end{itemize}
So, while one expects a general trend that operations with larger FLOP counts will take longer to execute, there is no expectation of smoothness.
Of particular note, the two operations with large spikes in runtime in most graphs are \emph{Fully Connected} layers, where the convolution is computed only once per image, with filters that are the size of the full input image.
Compared to other convolutions with similar FLOP counts, such operations offer less opportunity for parallelism and data reuse, and thus tend to be slower to execute.
However, these fully connected layers can be handled with a faster, less general version of convolution.
This special case is not fully implemented in Boda yet, and it appears cuDNN does not properly invoke its specialized version for these cases, perhaps since they are not explicitly marked as fully connected (though this can be easily deduced).
Thus, while the comparison for these points is (in some sense) fair, they should be considered somewhat as outliers in this analysis, and not given too much weight, pending further analysis and experimentation.
For the convenience of reviewers, we have also provided the list of operations as a table in Appendix A.

The NVIDIA GPU used is a Titan-X(Maxwell).
The AMD GPU used is an R9-Nano.
The Qualcomm GPU used is the Adreno 530 GPU portion of the Snapdragon 820 System-on-Chip (abbreviated ``SD820'' hereafter).
For the CUDA platform, we use the NVIDIA-provided nvrtc library to allow run-time compilation for CUDA.
All timings are performed using CUDA and OpenCL level timing functions, and thus should include only time spent on the GPU, and should not depend on the host CPU or other machine configuration details.
The input data given to the convolutions is all-non-zero pseudo-random noise.
Note that runtimes should not (in general) depend on the input data, as long as it has proper range and sparsity.
All outputs are cross-checked for numerical correctness using relative tolerances ranging from 0.001\% to 0.1\% depending on the operation.

\begin{figure*}
	\centering
	\includegraphics[width=1.0\textwidth]{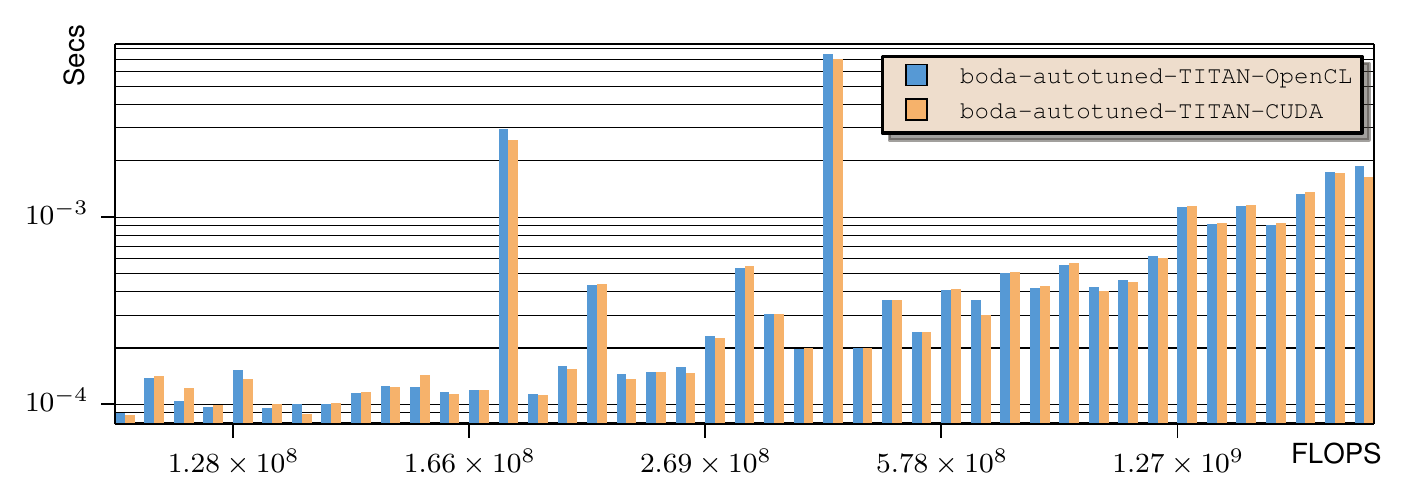}
	\caption{OpenCL vs CUDA. Runtime on NVIDIA Titan-X (Maxwell)}
	\label{fig:titan-ocl-vs-nvrtc}
\end{figure*}

\paragraph{Programming model portability -- OpenCL vs. CUDA.}
On NVIDIA hardware, we show that we can achieve almost identical per-operation runtime, using the same CUCL code, regardless of which programming interface we use (programming model portability).
This is (to the best of the author's knowledge) a novel illustration of the lack of importance of using CUDA versus OpenCL for a high-efficiency, difficult-to-implement GPU programming task.
Of course, this portability is only possible due to our framework level metaprogramming support.
Clearly, the framework support is critical because it allows for syntactic compatibility between the core languages of the two programming platforms.
Also, it abstracts away various higher-level issues in terms of compilation, allocation, scheduling, and execution that differ between the two platforms.
But more importantly, the framework shifts much of the implementation complexity into the metacode layer, which is relatively programming platform neutral.
Thus, the resulting OpenCL and CUDA code is quite simple and portable, using little beyond basic C constructs and the (common to OpenCL and CUDA) GPU threading model.
A comparison of CUDA vs. OpenCL efficiency on our benchmark set of operations is given in Figure \ref{fig:titan-ocl-vs-nvrtc}.
In the figure, all runtimes are for running each operation using the best function generated by Boda for that operation, selected by autotuning.
The two plotted cases differ only in the choice of backend (OpenCL or CUDA) for compilation and execution; the generated CUCL code for both cases is identical.
In both the the OpenCL and CUDA backends, it is possible to output the compiled ``binary'' code (in this case, NVIDIA PTX portable assembly code).
It can be noted that, for several cases that were inspected, the same CUCL source code yields the nearly the same PTX when compiled using either OpenCL or CUDA.
However, there are some minor differences: the addressing modes and internal LLVM compiler versions appear to slightly differ between NVIDIA's internal OpenCL and CUDA compilation flows.
These issues, combined with normal runtime variation/noise, can easily explain the remaining small differences in runtime between the OpenCL and CUDA cases.

\begin{figure*}
	\centering
	\includegraphics[width=1.0\textwidth]{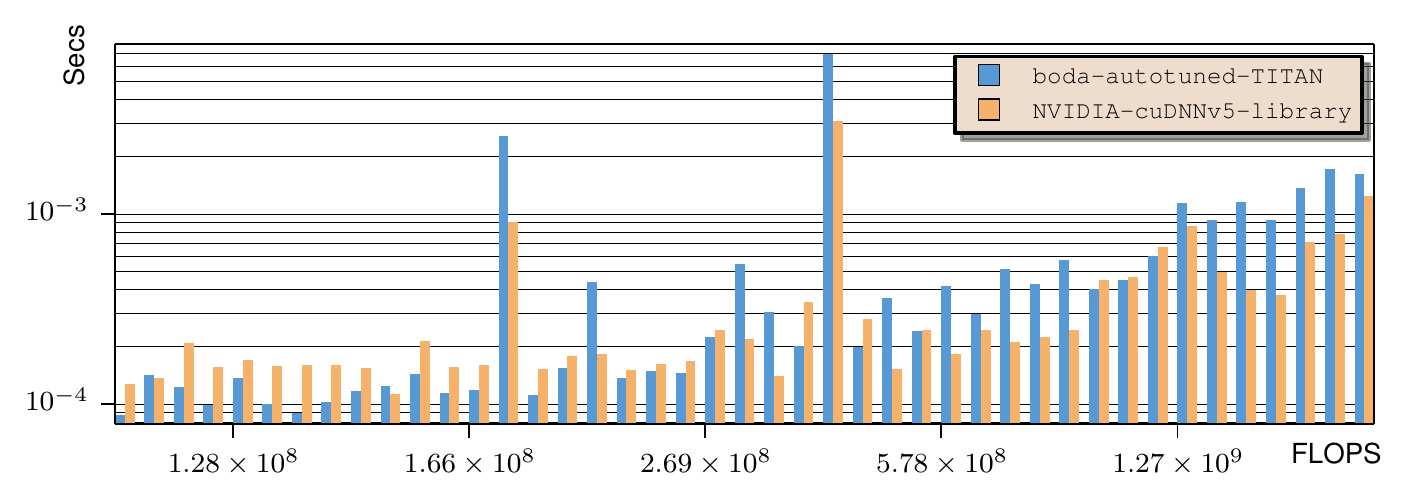}
	\caption{Comparison of Boda with cuDNNv5 on NVIDIA Titan-X}
	\label{fig:titan-tune}
\end{figure*}

As a gauge of the overall absolute quality of our results, in Figure \ref{fig:titan-tune}, we show the performance of Boda relative to the highly-tuned vendor CNN library cuDNN version 5.
Note that Boda is particularly slower in cases with 3x3 kernel sizes, where cuDNN is using Winograd convolution~\cite{lavin2015fast}, which is not yet implemented in Boda.
A case study to determine the effort/speed tradeoff of implementing Winograd convolution in Boda is a key topic of future work.
However, overall, we are reasonably competitive, and even faster than the reference library in a few cases.

\begin{figure*}
	\centering
	\includegraphics[width=1.0\textwidth]{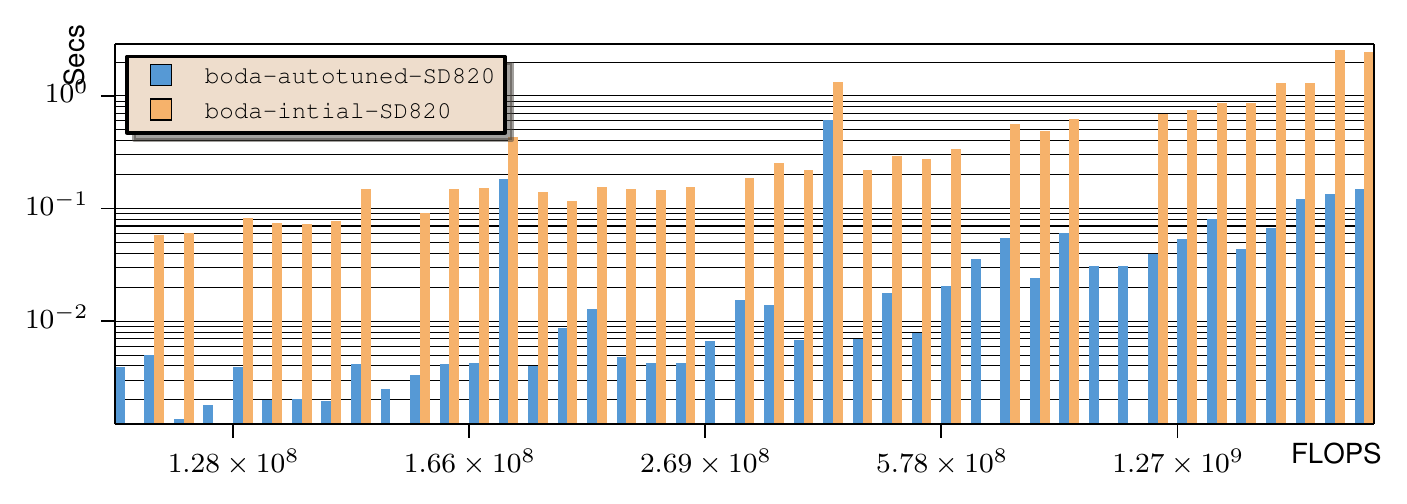}
	\caption{Initial vs. optimized results on Qualcomm Snapdragon 820}
	\label{fig:sd820-tune}
\end{figure*}

\paragraph{Tuning for Qualcomm Mobile GPUs .}
In Figure \ref{fig:sd820-tune}, the \emph{boda-initial} values show the initial (poor) performance when running the general-case fallback convolution variant on the SD820 platform.
When starting work on this platform, the general-case fallback variant was the only variant that could be run, since bugs in the Qualcomm OpenCL implementation and portability issues (primarily related to usage of shared memory and high register usage) prevented any of the existing optimized-for-NVIDIA variants from running at all.
The few missing bars in the \emph{boda-initial} series denote cases where even the simple fallback variant failed to compile or run.
However, with a few weeks of effort, we were able to create a new convolution variant that both worked around bugs in the Qualcomm platform as well as using some platform-tailored optimizations for memory access.
Additionally, based on analysis and experimentation, we added new points in the space of tuning parameters (specific thread and register blocking constants) to be searched over.
The final results of using the combination of the new variant and expanded tuning space are shown in the figure as \emph{boda-autotuned}, with the same meaning as in other figures: the values show the runtimes of the best variant and tuning parameters for each operation.

\begin{figure*}
	\centering
	\includegraphics[width=1.0\textwidth]{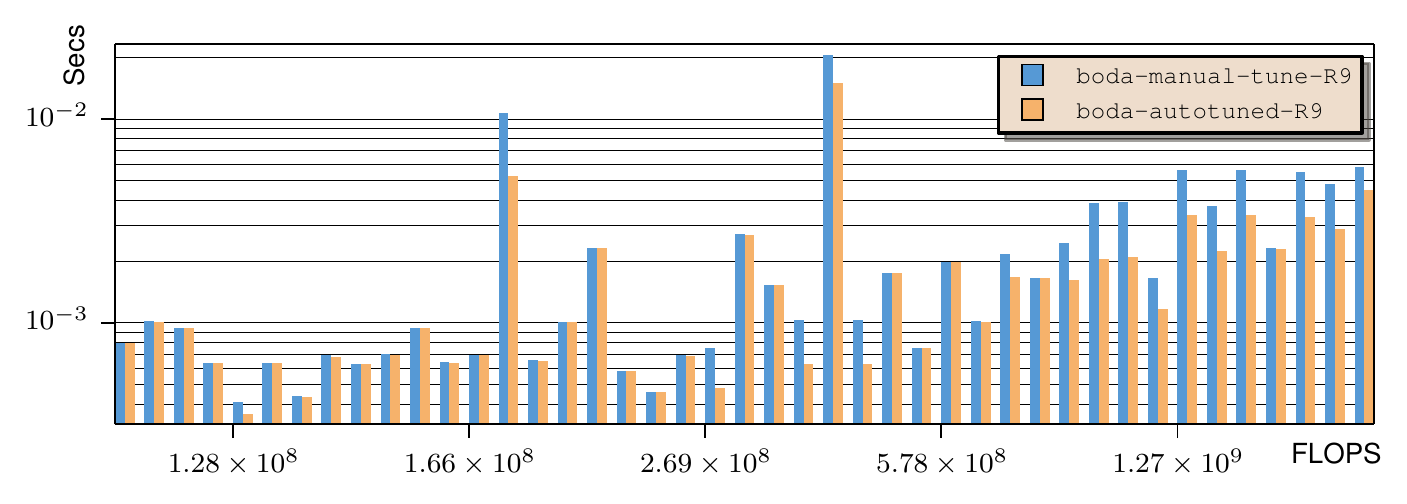}
	\caption{Manually-tuned and autotuned runtime on AMD R9-Nano (Fiji)}
	\label{fig:fiji-tune}
\end{figure*}

\paragraph{Improving efficiency by autotuning.}
We now move to some initial results on AMD hardware that demonstrate the value of autotuning.
Using the expanded library of variants and tuning space from targeting NVIDIA and Qualcomm hardware, we perform an experiment to isolate the effect of autotuning.
In Figure \ref{fig:fiji-tune}, we compare two cases.
First, we consider the runtimes one might achieve without autotuning.
In this case, it is too time consuming to select the best variant and tuning parameters for each operation individually.
Instead, the \emph{boda-manual-tune} values show the runtimes that result from:
\begin{itemize}
\item using a simple ``choose-most-specialized-possible'' heuristic to select the per-operation variant, and
\item choosing the \emph{single overall best} setting for tuning parameters, judged by the sum of runtime over all cases.
\end{itemize}
The second step in this process, while automatic, is designed to mimic the actual process and results of previous efforts at manual tuning that we performed prior to having autotuning support in our framework.
Thus, in addition to giving better results, autotuning requires \emph{much less} manual effort than manual tuning.
Additionally, the overall result of exploring the tuning space provides significant insight into this new platform.
By seeing which variants and tuning parameter settings work well, \emph{and which do not}, and comparing results across platforms, we can more quickly determine where to focus future optimization efforts.
As with all new platforms, it is difficult to predict how much speed improvement is possible with a given amount of optimization effort.
However, we are now well positioned to explore this question for the AMD platform as future work.

\begin{figure*}
	\centering
	\includegraphics[width=1.0\textwidth]{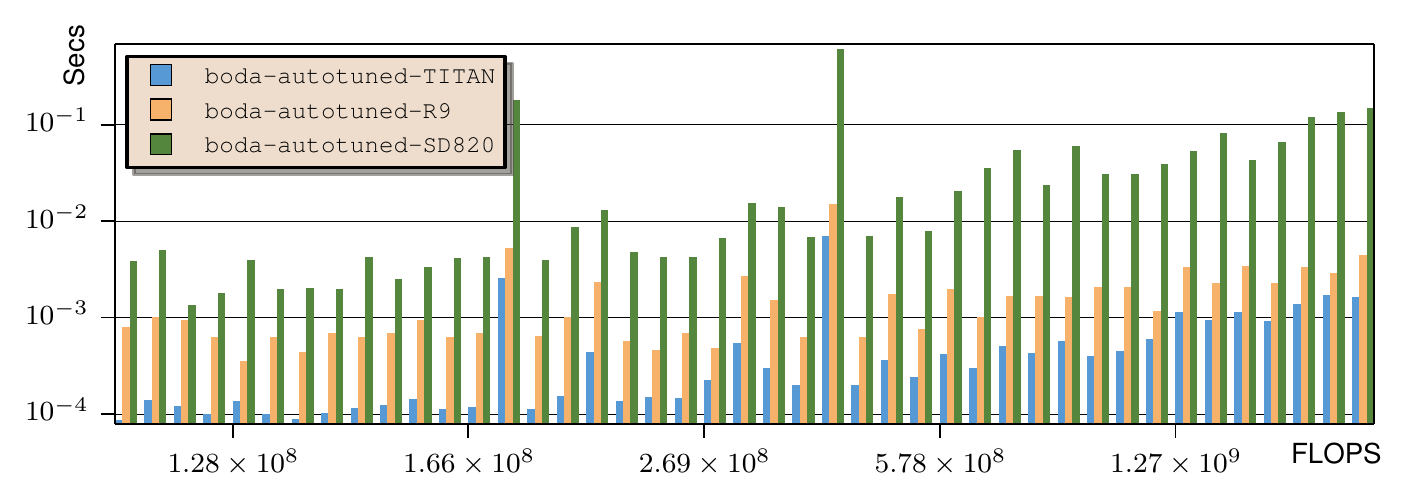}
	\caption{Autotuned runtime on NVIDIA Titan-X, AMD R9-Nano, and Qualcomm Snapdragon 820}
	\label{fig:all-plats}
\end{figure*}

\paragraph{Performance portability on different targets.}
In Figure \ref{fig:all-plats}, we show the overall portability of our benchmark convolution operations across three different platforms.
Using a single framework and library of variants and tuning parameters, we achieve reasonable performance across three different hardware platforms (AMD, NVidia, and Qualcomm) and different two programming platforms (OpenCL and CUDA).
Note that the generated code has no dependencies on any platform-specific libraries (or any libraries at all), and all code is generated and compiled at run-time specific to each operation instance.
In particular, for testing, the framework can run the same operation on all platforms supported \emph{within a single process} and compare full results across platforms on the fly.
Currently, the results for the AMD platform are significantly slower than those on the NVIDIA platform, especially for the smaller (lower FLOP count) operations.
However, bear in mind that these are initial results, and at most show that OpenCL lacks performance portability even across relatively similar platforms with comparable peak computational and memory performance.
Similarly, while the SD820 results are much slower than the NVIDIA results (by perhaps 2 orders of magnitude), it must be remembered that the SD820 GPU is (by design) a much smaller device with much lower power usage and correspondingly lower peak performance.
At this time, we present these results mainly to show our portability, and not to directly compare these platforms.
However, with modest additional optimization efforts on the AMD and Qualcomm platforms, one may be able to draw fairer comparisons between these disparate platforms.


%% file: 07_relatedworks.tex
\section{Related Works}
\label{sec:relatedworks}
\vsp

Early NN frameworks such cuda-convnet~\cite{alexnet} and Caffe~\cite{jia2014caffe} performed CNN convolutions by leveraging Nvidia’s cuBLAS~\cite{cuBLAS} matrix math (BLAS) library.
However, there are various limitations to the BLAS-based approach to NN convolution.
In particular, it:
\begin{itemize}
\item does not reuse data between spatially overlapping input windows,
\item sometimes requires expensive input and output transformations to convert 4D-Arrays into 2D matrices,
\item and does not allow fusion of an activation function with the convolution operation.
\end{itemize}
Additionally, the underlying matrix-matrix multiply function may not be well optimized for the problem sizes required.
Finally, other higher-level optimizations, such as Winograd convolution~\cite{lavin2015fast}, cannot leverage existing BLAS kernels.

In any event, various purpose-built libraries to perform NN convolution have improved speed and efficiency over BLAS-based approaches.
NVIDIA's cuDNN~\cite{cuDNN} library is the most common and popular of such libraries.
The library exposes an API for directly performing convolutions using a variety of algorithms.
While this achieves much higher efficiency than BLAS-based approaches~\cite{soumith-bench}, the library is closed-source and limited to NVIDIA hardware.
Thus, it may not be extended by the community to support less-common or new operations or to target new hardware platforms.

In parallel with the closed development of cuDNN, a more open family of libraries based on an assembly-language-level metaprogramming flow~\cite{maxDNN}~\cite{neon} was developed.
The performance of this approach, as embodied in Nervana System's ``neon'' framework, generally was similar or better than that of cuDNN at any given point in time.
However, as with cuDNN, this approach is limited to NVIDIA hardware.
Further, the use of perl-based metaprogramming to generate low-level GPU assembly code creates significant hurdles to extending this approach for new operations or targeting multiple platforms.
We operate instead at the higher abstraction level of CUCL, and use a C++-hosted string-template based metaprogramming approach.
We argue that our approach of writing C++ code that generates C code is relatively easier to work with and extend than writing perl to generate assembly.
In particular, many constructs look roughly the same at the metacode and code levels.
As shown in the example in Section~\ref{sec:approach}, to statically unroll a loop, one simply moves the loop from the code to the metacode, and ``escapes'' the body of the loop to print the code it previously contained.
In essence, we claim the similarity and compatibility between the metacode and code languages eases the burden on the programmer to operate across both levels.
Further, rather than simply creating a convolution library, we span the entire flow from compute graph to execution, which allows for additional freedom and optimizations.

One common approach to metaprogramming is to use built-in language level metaprogramming facilities.
In particular, C++ templates are commonly used for high performance GPU metaprogramming with CUDA.
However, C++ templates have the following disadvantages as compared with this work:
\begin{itemize}
\item C++ template support for OpenCL is only starting to become available.
\item Like perl, C++ templates are a significantly different language compared to C, and are generally considered difficult to use.
\item C++ templates do not offer the practical ability to perform complex, significant meta-level functions (such as running compile-time search algorithms, using complex nested conditionals, or easily reusing generator sub-functions), which can be an important ``escape-hatch'' when needing full control over generated code sequences.
\item C++ templates do not allow the ability to inspect the generated C level code for a given instantiation for debugging and analysis.
\end{itemize}


More closely related to this work are two projects that derive from early efforts to add OpenCL support to the Caffe framework: Greentea LibDNN~\cite{tschopp2015efficient} and cltorch~\cite{perkins2016cltorch}.
Both projects originally used BLAS-based approaches, but both have moved toward metaprogramming-based special-purpose code generation for DNN operations.
However, based on published results~\cite{perkins2016cltorch}~\cite{soumith-bench}, cltorch and Greentea do not appear to be currently competitive with cuDNN on NVIDIA platforms (unlike this work).
While these results are likely out of date, the limited documentation make independent benchmarking of the current state of the projects problematic.
Finally, it appears that neither project currently supports Qualcomm GPUs.

On the topic of programming model portability, any comparison must consider both performance portability and programming model portability at the same time, which requires a common benchmarking methodology.
The OpenCL-based cltorch project also provides a comparison with a similar CUDA based approach (CUDA torch).
However, being separate projects, this comparison does not imply programming model portability for cltorch -- nor is the speed of cltorch and CUDA torch particularly close~\cite{perkins2016cltorch}.

Overall, direct per-operation speed comparison between Greentea, cltorch, and this work seems currently difficult to achieve, but is a good topic for future work and/or collaboration to create a unified benchmarking environment to help clarify these issues~\cite{deepmark}.


%% file: 08_conclusion.tex
\section{Conclusions}
\label{sec:conclusions}
\vsp

Boda is a new framework for rapid prototyping and productive development of efficient GPU code for DNN operations.
In particular, it supports metaprogramming and autotuning with key features that enable both programming model portability and performance portability. 
Experimental results show that Boda’s variant selection and autotuning support eases the path to portable, efficient implementations.
On NVIDIA hardware, we achieve performance competitive with the vendor library, and we can achieve that performance using either the OpenCL or CUDA programming model.
On Qualcomm hardware, we show that we can quickly develop new variants and otherwise tune our generated code to achieve reasonable performance on a mobile GPU.
On AMD hardware, we show that autotuning and profiling pre-existing code on a new platform provides a good foundation for platform-specific optimization efforts as future work.
Further, as an open, vendor-neutral framework, we avoid dependencies on any specific hardware platforms or unextensible vendor libraries.
Thus, our framework provides a productive method for implementing existing and new DNN operations while targeting various hardware platforms.

%% file: 09_appendix.tex
\section*{Appendix A}
\label{sec:appendix}
\vsp

\subsection*{Benchmark Operations List (to be omitted for publication; will be available online)}
\label{app:ops-list}

Column meanings:
\begin{itemize}
\item KSZ: kernel size; all kernels are square, so the value given is the kernel size in both X and Y
\item Stride: uniform in spatial dimensions, so the value given is the stride in both X and Y
\item OC: number of output channels, redundant with $chan$ dimension of output
\item $in$ and $out$: 4D-Array sizes of input and output as $y \dx x \dx chan$
\item FLOPs: per-operation FLOP count
\end{itemize}
Note: the two operations with 4096 output channels are the fully-connected-layer convolutions mentioned as outliers in the results section.

\begin{longtable}{lllllll}
\hline
KSZ  & S & OC & $B$ & $in$ & $out$ & FLOPs \\ 
\hline
\endhead
\input{figs/ops-out-sort-by-flops.tex}
\label{tab:cnn-op-info}
\end{longtable}


%% file: figs/ops-out-sort-by-flops.tex
5 & 1 & 32 & 5 & $  28 \dx 28 \dx 16 $ & $  28 \dx 28 \dx 32 $ &  1.00352e+08 \\ 
5 & 1 & 64 & 5 & $  14 \dx 14 \dx 32 $ & $  14 \dx 14 \dx 64 $ &  1.00352e+08 \\ 
1 & 1 & 256 & 5 & $  7 \dx 7 \dx 832 $ & $  7 \dx 7 \dx 256 $ &  1.04366e+08 \\ 
1 & 1 & 112 & 5 & $  14 \dx 14 \dx 512 $ & $  14 \dx 14 \dx 112 $ &  1.12394e+08 \\ 
1 & 1 & 128 & 5 & $  14 \dx 14 \dx 512 $ & $  14 \dx 14 \dx 128 $ &  1.28451e+08 \\ 
1 & 1 & 64 & 5 & $  28 \dx 28 \dx 256 $ & $  28 \dx 28 \dx 64 $ &  1.28451e+08 \\ 
1 & 1 & 64 & 5 & $  56 \dx 56 \dx 64 $ & $  56 \dx 56 \dx 64 $ &  1.28451e+08 \\ 
1 & 1 & 128 & 5 & $  14 \dx 14 \dx 528 $ & $  14 \dx 14 \dx 128 $ &  1.32465e+08 \\ 
1 & 1 & 144 & 5 & $  14 \dx 14 \dx 512 $ & $  14 \dx 14 \dx 144 $ &  1.44507e+08 \\ 
1 & 1 & 96 & 5 & $  28 \dx 28 \dx 192 $ & $  28 \dx 28 \dx 96 $ &  1.44507e+08 \\ 
1 & 1 & 384 & 5 & $  7 \dx 7 \dx 832 $ & $  7 \dx 7 \dx 384 $ &  1.56549e+08 \\ 
1 & 1 & 160 & 5 & $  14 \dx 14 \dx 512 $ & $  14 \dx 14 \dx 160 $ &  1.60563e+08 \\ 
1 & 1 & 160 & 5 & $  14 \dx 14 \dx 528 $ & $  14 \dx 14 \dx 160 $ &  1.65581e+08 \\ 
1 & 1 & 4096 & 5 & $  1 \dx 1 \dx 4096 $ & $  1 \dx 1 \dx 4096 $ &  1.67772e+08 \\ 
1 & 1 & 192 & 5 & $  14 \dx 14 \dx 480 $ & $  14 \dx 14 \dx 192 $ &  1.80634e+08 \\ 
5 & 1 & 128 & 5 & $  14 \dx 14 \dx 32 $ & $  14 \dx 14 \dx 128 $ &  2.00704e+08 \\ 
3 & 1 & 320 & 5 & $  7 \dx 7 \dx 160 $ & $  7 \dx 7 \dx 320 $ &  2.25792e+08 \\ 
1 & 1 & 384 & 5 & $  13 \dx 13 \dx 384 $ & $  13 \dx 13 \dx 384 $ &  2.49201e+08 \\ 
1 & 1 & 128 & 5 & $  28 \dx 28 \dx 256 $ & $  28 \dx 28 \dx 128 $ &  2.56901e+08 \\ 
1 & 1 & 256 & 5 & $  14 \dx 14 \dx 528 $ & $  14 \dx 14 \dx 256 $ &  2.64929e+08 \\ 
1 & 1 & 96 & 5 & $  54 \dx 54 \dx 96 $ & $  54 \dx 54 \dx 96 $ &  2.68739e+08 \\ 
3 & 1 & 384 & 5 & $  7 \dx 7 \dx 192 $ & $  7 \dx 7 \dx 384 $ &  3.2514e+08 \\ 
3 & 1 & 208 & 5 & $  14 \dx 14 \dx 96 $ & $  14 \dx 14 \dx 208 $ &  3.52236e+08 \\ 
1 & 1 & 1000 & 5 & $  6 \dx 6 \dx 1024 $ & $  6 \dx 6 \dx 1000 $ &  3.6864e+08 \\ 
1 & 1 & 1024 & 5 & $  6 \dx 6 \dx 1024 $ & $  6 \dx 6 \dx 1024 $ &  3.77487e+08 \\ 
6 & 1 & 4096 & 5 & $  6 \dx 6 \dx 256 $ & $  1 \dx 1 \dx 4096 $ &  3.77487e+08 \\ 
3 & 1 & 224 & 5 & $  14 \dx 14 \dx 112 $ & $  14 \dx 14 \dx 224 $ &  4.42552e+08 \\ 
1 & 1 & 256 & 5 & $  27 \dx 27 \dx 256 $ & $  27 \dx 27 \dx 256 $ &  4.77757e+08 \\ 
3 & 1 & 256 & 5 & $  14 \dx 14 \dx 128 $ & $  14 \dx 14 \dx 256 $ &  5.78028e+08 \\ 
5 & 1 & 96 & 5 & $  28 \dx 28 \dx 32 $ & $  28 \dx 28 \dx 96 $ &  6.02112e+08 \\ 
3 & 1 & 288 & 5 & $  14 \dx 14 \dx 144 $ & $  14 \dx 14 \dx 288 $ &  7.31566e+08 \\ 
3 & 1 & 128 & 5 & $  28 \dx 28 \dx 96 $ & $  28 \dx 28 \dx 128 $ &  8.67041e+08 \\ 
3 & 1 & 320 & 5 & $  14 \dx 14 \dx 160 $ & $  14 \dx 14 \dx 320 $ &  9.03168e+08 \\ 
11 & 4 & 96 & 5 & $  224 \dx 224 \dx 3 $ & $  54 \dx 54 \dx 96 $ &  1.01617e+09 \\ 
11 & 4 & 96 & 5 & $  227 \dx 227 \dx 3 $ & $  55 \dx 55 \dx 96 $ &  1.05415e+09 \\ 
7 & 2 & 64 & 5 & $  224 \dx 224 \dx 3 $ & $  112 \dx 112 \dx 64 $ &  1.18014e+09 \\ 
3 & 1 & 1024 & 5 & $  6 \dx 6 \dx 384 $ & $  6 \dx 6 \dx 1024 $ &  1.27402e+09 \\ 
3 & 1 & 256 & 5 & $  13 \dx 13 \dx 384 $ & $  13 \dx 13 \dx 256 $ &  1.4952e+09 \\ 
3 & 1 & 384 & 5 & $  13 \dx 13 \dx 256 $ & $  13 \dx 13 \dx 384 $ &  1.4952e+09 \\ 
3 & 1 & 192 & 5 & $  28 \dx 28 \dx 128 $ & $  28 \dx 28 \dx 192 $ &  1.73408e+09 \\ 
3 & 1 & 384 & 5 & $  13 \dx 13 \dx 384 $ & $  13 \dx 13 \dx 384 $ &  2.24281e+09 \\ 
3 & 1 & 192 & 5 & $  56 \dx 56 \dx 64 $ & $  56 \dx 56 \dx 192 $ &  3.46817e+09 \\ 
5 & 1 & 256 & 5 & $  27 \dx 27 \dx 96 $ & $  27 \dx 27 \dx 256 $ &  4.47898e+09 \\ 